\documentclass[a4paper,conference]{IEEEtran}

\usepackage{cite}

\ifCLASSINFOpdf
  \usepackage[pdftex]{graphicx}
\else
\fi
\usepackage{amsmath}
\usepackage{xcolor}
\usepackage{textcomp}
\usepackage{hyperref}
\hypersetup{colorlinks=true, urlbordercolor={1 1 1}, linkcolor={black}, raiselinks=false}
\usepackage{bbm} 
\newcommand{\norm}[1]{\|#1\|} 
\usepackage{algorithm}
\usepackage{algpseudocode}
\usepackage{float}
\usepackage{booktabs} 

\begin{document}
%
\title{An Empirical Analysis of Visual Features for Multiple Object Tracking in Urban Scenes}

\author{\IEEEauthorblockN{Mehdi Miah, Justine Pepin, Nicolas Saunier and Guillaume-Alexandre Bilodeau}
\IEEEauthorblockA{Polytechnique Montr\'{e}al \\ Montr\'{e}al, Canada}}


%


\maketitle

\begin{abstract}
This paper addresses the problem of selecting appearance features for multiple object tracking (MOT) in urban scenes. Over the years, a large number of features has been used for MOT. However, it is not clear whether some of them are better than others. Commonly used features are color histograms, histograms of oriented gradients, deep features from convolutional neural networks and re-identification (ReID) features. In this study, we assess how good these features are at discriminating objects enclosed by a bounding box in urban scene tracking scenarios. Several affinity measures, namely the $\mathrm{L}_1$, $\mathrm{L}_2$ and the Bhattacharyya distances, Rank-1 counts and the cosine similarity, are also assessed for their impact on the discriminative power of the features.
Results on several datasets show that features from ReID networks are the best for discriminating instances from one another regardless of the quality of the detector. If a ReID model is not available, color histograms may be selected if the detector has a good recall and there are few occlusions; otherwise, deep features are more robust to detectors with lower recall.
The project page is \url{www.mehdimiah.com/visual_features}.
\end{abstract}


%
\IEEEpeerreviewmaketitle

\section{Introduction}

    Cities are faced with many challenges, including how to move people safely and efficiently for their daily activities. Data on the movement of all road users is therefore necessary. Such data can be collected automatically through various kinds of sensors, including video cameras with computer vision algorithms. The main task is to detect and track all roads users, which is also called multiple object tracking (MOT). This is one example, among many, of the use of MOT.

   Many state-of-the-art MOT methods rely on the strategy called ``tracking-by-detection''~\cite{luo2017MultipleObject}: first, they detect objects of interest, such as vehicles or pedestrians, then they link the detections between frames to create trajectories. For the second step, various features are used: appearance, spatial information and motion~\cite{gladh2016DeepMotion}. Even if MOT is a well-studied problem~\cite{ooi2018MultipleObject,jodoin2016TrackingAll,yang2017MultipleObject,leal-taixe2015MOTChallenge2015}, there are still many unsolved challenges limiting the quality of the results. One of them is describing objects' appearance. It should be possible to distinguish every tracked object from the others, while at the same time considering that the appearance of an object might change over time because of a viewpoint change and illumination variations. Therefore, selecting the most discriminative features and finding a proper way to compare them become two key elements in the process of visual appearance modeling. 
    
    Should handcrafted features be used, or should the object appearance be learned? Given the fact that in MOT, several aspects, such as appearance, spatial and motion information, are usually investigated at the same time, it is difficult to tell whether a method is better because of the chosen feature, or the data association method, or the method to predict where the object should be in the future. 
    
    Recently, Kornblith et al.~\cite{kornblith2019BetterImageNet} studied how models with high performance on ImageNet~\cite{deng2009ImageNetlargescale,russakovsky2015ImageNetLarge} actually performed for classification on other datasets. The answer is comforting: the better the models are on ImageNet, the better they are on other datasets. However, can the same conclusion be drawn on a downstream task such as tracking? Indeed, MOT encounters atypical challenges such as the need for instance classification, deformations, illumination changes, occlusions, blur, etc. Some of these challenges are absent from the ImageNet dataset. Moreover, the classification required in tracking is more fine-grained to distinguish all object instances: models trained on ImageNet succeed when they correctly classify persons as persons whereas for the MOT task, these persons must be discriminated from one another.
    
    In this paper, we assess the performance of popular visual features to describe objects in MOT in various urban scene scenarios. These are among the most popular scenarios in MOT, and the focus of the most popular MOT datasets. Therefore, the objects of interest to describe are mostly pedestrians and various vehicles. To avoid interference from other MOT components, we only focus on the visual appearance description and comparison for image regions enclosed by bounding boxes (BB). No spatial or motion information are used in this paper. 
    The results suggest that re-identification (ReID) features are the best visual features for MOT tasks. When these features are not available, deep features may be used and give better performance than the color histogram when objects are further apart in time. The HOG features critically degrade when the detector provides imprecise BBs. 

    
    The main contributions of this paper are: 
    \begin{itemize}
        \item a comparison of visual descriptors on four MOT datasets;
        \item a new methodology to compare features for the MOT task; 
        \item an analysis of descriptors and affinity measures performance according to the size of objects, the precision of the bounding boxes  and the elapsed time between observations.
    \end{itemize}


\section{Related Work} 
\label{rwork}

    A large variety of appearance features have been used for tracking. Some of them are briefly reviewed in this section.
    
    One of the most popular features for MOT is the color histogram. Among others, color histograms were used in the work of~\cite{riahi2015Multipleobject, dianzhu2012realtimerobust, sun2013Multiplepedestrians}. In the work of Riahi et al.~\cite{riahi2015Multipleobject}, color histograms are combined with other features such as optical flow and a sparse representation. Optical flow calculates the motion vector of pixels between two frames, while a sparse representation reconstructs an image region using templates from the image regions of a model object and trivial templates, which contain only one non-zero value. If the visual appearance of an object is different from the model object, the reconstruction will require many trivial templates. In the case of the work of Zhu et al.~\cite{dianzhu2012realtimerobust} and Sun et al.~\cite{sun2013Multiplepedestrians}, color histograms are combined with histograms of oriented gradients (HOG). 
    
    
    While the color histogram focuses on the general color appearance of an object, HOG focuses on the texture of an object (spatial arrangement of the colors). Because HOG features are calculated using gradient magnitudes as weights, they often also capture the general shape of an object. Therefore an HOG feature can be seen both as a texture and a shape descriptor. The MOT methods presented in~\cite{dianzhu2012realtimerobust,heimbach2017Resolvingocclusion,sun2013Multiplepedestrians} for example rely on HOG. In the work of Heimbach et al.~\cite{heimbach2017Resolvingocclusion}, HOG is used solely in combination with a Kalman filter to predict object position. 
    
    
    Many works use deep features as universal descriptors for MOT~\cite{wang2014Learningdeep,sadeghian2017TrackingUntrackable, tang2017MultiplePeople, ma2015HierarchicalConvolutional, danelljan2016CorrelationFilters}. The object appearance is described with features from VGG-16 in both the work of Tang et al.~\cite{tang2017MultiplePeople} and Sadeghian et al.~\cite{sadeghian2017TrackingUntrackable}, while the work of Wang et al.~\cite{wang2014Learningdeep} uses a two-layer custom Convolutional Neural Network (CNN). Class labels (e.g.\ car, pedestrian, bike) were also used recently as a coarse description of an object appearance~\cite{ooi2018MultipleObject}. As for~\cite{ma2015HierarchicalConvolutional}, the authors worked with VGG-19 from which multiple outputs from different layers were extracted. Finally, recent works include ReID features~\cite{chen2017TripletLoss,wojke2017Simpleonline,zhou2019OmniScaleFeature,zhou2019LearningGeneralisable}. These features are computed by learning a model to predict if two detections from two points of view are instances of the same object.
    
    
    Surprisingly, we could only find one work that compared features for MOT~\cite{gil1996Comparingfeatures}. Because it dates back to 1996, the features that were compared are the distance between the center of gravity, the size of the BBs, and correlation between object pixels.  
    
    
        
    
\section{Tested visual features, affinity measures and datasets}
\label{perimeter}
\begin{figure*}[ht]
    \centering               
    \includegraphics[scale = 0.45, trim={20 60 60 60},clip]{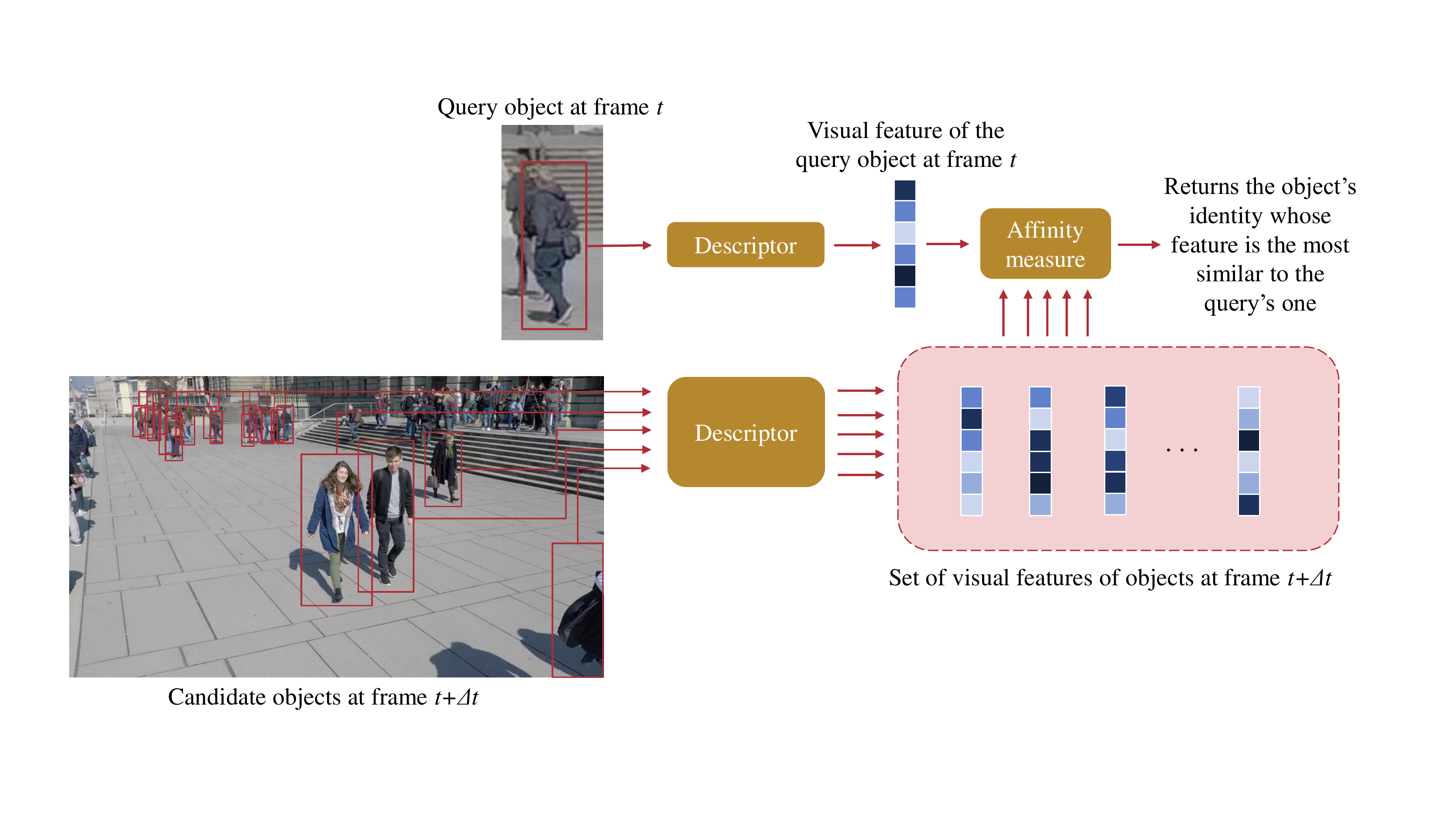}
    \caption{High-level explanation of our experimental methodology. From bounding boxes, a feature descriptor is calculated for both a query object and candidate matching objects in another frame. Then, the affinity measure is calculated for all query and candidate pairs, and the best match is returned and evaluated based on the ground truth.}
    \label{fig:explanation}
    \end{figure*}

Figure~\ref{fig:explanation} gives an overview of our visual feature evaluation strategy. Given a bounding box-enclosed object extracted from a frame, it is first described with a feature descriptor. Then, we select another frame where this object is present, describe all objects in this frame with the same feature descriptor, and compute an affinity measure to select the most similar object in term of visual appearance. In the following, we describe the visual features, affinity measures and datasets selected for our evaluation. Given the available MOT datasets, we assume that objects are either pedestrians or vehicles.

\subsection{Visual features}

    There are many ways to obtain a description (a numerical vector) from an image of an object enclosed by a BB. We selected eight popular visual descriptors among four categories: color histograms-based, gradient-based, CNN-based and ReID-based models.


\paragraph{Color and grayscale histograms}
    the RGB color histogram descriptor consists in counting the number of occurrences of each color inside a BB. We elected to use quantified histograms because they are more resilient to noise. For grayscale 1D-histogram, we used 32 bins. For color histogram, similarly, each channel of the image gives an histogram; after concatenation, we obtain a vector of size 96.

\paragraph{Histograms of oriented gradients (HOG)}
    contrarily to color-based histograms, gradient-based descriptors are more robust to illumination change. HOG~\cite{dalal2005Histogramsoriented} is one famous example. It captures texture information in the image as well as shape information about the object. To obtain an HOG descriptor, the image is first convolved with kernels to extract vertical and horizontal gradients. From these gradients, their angles and magnitudes can be obtained. The angles are typically quantified between 0 and 180\textdegree, as it was shown experimentally that ignoring the sign of the angle gives better results. Histograms are then constructed for several overlapping cells by counting the occurrence of quantified angles weighted by their gradient magnitudes. Since in our experimental setup, the datasets contain either vehicles or pedestrians, the HOG vectors are the same size for all candidate objects in each dataset : the BB is resized to $64\times 128$ pixels for pedestrians and $64 \times 64$ for vehicles, with cells of size $8\times 8$ and blocks of size $16 \times 16$.
    By quantifying angles into nine bins every 20\textdegree, this results in a vector of 1764 elements for vehicles and 3528 for pedestrians.
    


\paragraph{CNN-based features}
    since their breakthrough in 2012~\cite{krizhevsky2012ImageNetClassification}, CNNs are commonly used in computer vision for classification tasks. We used four different architectures which have great performance on ImageNet to extract visual features. For each network, we removed the last fully-connected layer to obtain a descriptor~\cite{donahue2014DeCAFDeep}. We evaluated VGG-19~\cite{simonyan2015VeryDeep}, ResNet-18~\cite{he2016DeepResidual}, DenseNet-121~\cite{huang2017DenselyConnected} and EfficientNet-B0~\cite{tan2019EfficientNetRethinking} architectures, respectively providing a description vector of size 4096, 512, 1024 and 1280, having respectively 140, 11, 7 and 4 millions parameters and resulting ImageNet top-1 error rates of respectively 27.6\%, 30.2\%, 25.4\% and 23.4\%.
    
\paragraph{Re-identification network}
    in the case of pedestrian tracking, we evaluated a re-identification method named OSNet-AIN~\cite{zhou2019LearningGeneralisable} containing 3 millions parameters giving ReID features of size 512. For vehicles, we extracted ReID features from the model of~\cite{wu2018VehicleReIdentification} containing 24 millions parameters providing vectors of size 2048.
    


\subsection{Descriptor affinity measures}
Each object BB in a frame is described with a numerical vector $\mathbf{x}$. So, given another image containing $m$ object BBs, each of them described by a vector $\mathbf{y}_j$, the aim is now to find the ``most similar'' vector to $\mathbf{x}$ among $\{\mathbf{y}_j | j \in [\![1,m]\!] \}$ hoping that the two vectors are instances of the same object.

The following five different affinity measures were used to compare the visual feature vectors.

\paragraph{$\mathrm{L}_1$ and $\mathrm{L}_2$ distances}  
    two common ways to compute affinity between vectors are the $\mathrm{L}_1$ and $\mathrm{L}_2$ distances. Given two vectors $\mathbf{x}$ and $\mathbf{y}$, for $p \geq 1 $, the $\mathrm{L}_p$ distance is given by
    
    \begin{equation}
        \mathrm{L}_p(\mathbf{x}, \mathbf{y})= \left(\sum_{i=1}^n |x_i-y_i|^p \right)^{\frac{1}{p}}
    \end{equation}
    where  $i$ refers to the $i^{th}$ element of each vector and $n$ is the length of the vectors to compare. The smaller the distance is, the more similar the vectors are. 
    

\paragraph{Rank-1 counts} 
    to compare feature vectors from a CNN, the Rank-1 counts was proposed in~\cite{jin2017EndToEndFace}. It was shown to be efficient to compare deep features. It works by comparing a pair of vectors ($\textbf{x}, \textbf{y}$) to other possible pairs ($\textbf{x}, \textbf{z}$) such that $\textbf{z} \neq \textbf{y}$. The underlying principle is to find the vector whose elements are closer to a query vector. It is computed as follows: 
    
    \begin{equation}
        C_{\text{rank}1}(\textbf{x}, \textbf{y})=\sum_{i=1}^n  1\Big( |x_i-y_i|< \min_{\textbf{z}, \textbf{z} \neq \textbf{y}} |x_i-z_i|\Big)
    \end{equation}
    where $1$ is an indicator function that takes value 1 if the expression in the argument is true and 0 otherwise. The expression verifies whether the $i^{th}$ element of $\textbf{x}$ is strictly closer to the corresponding element of $\textbf{y}$ compared to all other candidate vectors $\textbf{z}$. The larger the $C_{\text{rank}1}$ is, the more similar the objects are. 
    
\paragraph{Bhattacharyya distance}
this distance~\cite{bhattacharyya1943measure} measures the affinity between two distributions as follows: 

   \begin{equation}
        D_B(\textbf{x}, \textbf{y}) = - \ln(\sqrt{\textbf{x}}^\top \sqrt{\textbf{y}})
    \end{equation}
The smaller the $D_B$ is, the more similar the objects are.

\paragraph{Cosine similarity}
for two vectors $\textbf{x}$ and $\textbf{y}$, the cosine similarity is given by : 

   \begin{equation}
        S_C(\textbf{x}, \textbf{y}) = \dfrac{\textbf{x}^{\top}\textbf{y}}{\norm{\textbf{x}} \norm{\textbf{y}}}
    \end{equation}
    
The smaller the $S_C$ is, the less similar the objects are.

\subsection{Datasets}
We tested the visual features on four datasets commonly used in MOT. Two of them focus on pedestrians and two others on vehicles. Table~\ref{table:datasets} summarizes some of their characteristics.

    \begin{table}[ht]
    \caption{Dataset statistics: FPS: framerate, $\#S$: number of sequences, $\bar{F}$: average number of frames per sequence, $\bar{P}$: average number of pedestrians per frame, $\bar{V}$: average number of vehicles per frame and $\bar{S}$: average object size}
    \label{table:datasets}
    \centering
    \begin{tabular}{lcccccc}
     \toprule
     Name & FPS & $\#S$ & $\bar{F}$ & $\bar{P}$ & $\bar{V}$ & $\bar{S}$ \\
     \hline
     WildTrack~\cite{chavdarova2018WILDTRACKMultiCamera} & 2 & 7 & 400 & 15 & 0 & 85x286\\
     MOT17~\cite{leal-taixe2015MOTChallenge2015}& 14/25/30 & 7 & 759 & 21 & 0 &  85x230\\
     DETRAC~\cite{wen2020UADETRACnew} & 25 & 60 & 1368 & 0 & 7 & 97x66\\
     UAVDT~\cite{du2018UnmannedAerial} & 30 & 50 & 808 & 0 & 20& 39x34\\
     \bottomrule
    \end{tabular}
    \end{table}

\section{Experimental methodology}
\label{experiments}
    
    To evaluate the performance of 35 descriptor-affinity pairs (each pair composed of a feature and an affinity measure except pairs between a non histogram-based feature and the Bhattacharyya distance), we tried to link two bounding boxes referring to the same object throughout a video. For that, given a BB-enclosed object extracted from a frame, we described it with a feature descriptor. Then we select another frame where this object is present, described all objects in this frame with the same feature descriptor, and compute an affinity measure to select the most similar object to the one in the first frame (Figure \ref{fig:explanation}). We then verify if the match is correct based on the ground truth. 
    
    \subsection{Data preparation}
    It should be noted that working with the true BBs is a too ideal scenario. In practice, when ``tracking-by-detection'' is applied, the detection algorithm may omit an object, detect non-object elements or the predicted BBs may be slightly shifted. In order to simulate the BBs returned by a detector, we introduced noise in two ways. 
    
\begin{figure}[ht]
    \centering               
    \includegraphics[scale = 0.27]{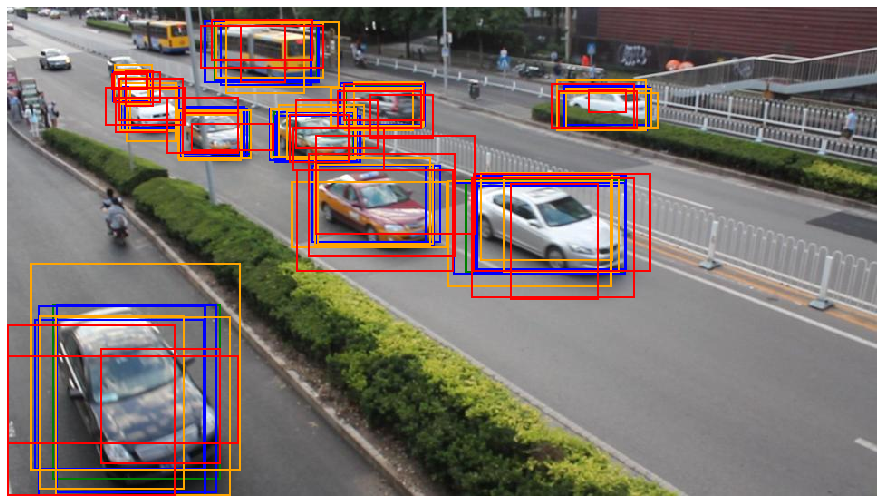}
    \caption{Examples of noisy bounding boxes on a frame of the DETRAC dataset: for each object of interest, three examples for each $\sigma$ are displayed to illustrate the variability of noisy BBs. The color code is as follows: green for $\sigma=0$ (the ground truth BB), blue for $\sigma=0.05$, orange for $\sigma = 0.1$ and red for $\sigma = 0.2$. Best viewed in color.}
    \label{fig:sigma}
    \end{figure}

    \paragraph{Noisy coordinates}
    the first way is to add a white Gaussian noise to each coordinate independently. Given a BB $(x_m, y_m, x_M, y_M)$ where $(x_m, y_m)$ is the top-left coordinate of the BB and  $(x_M, y_M)$ the bottom-right one, noisy coordinates are obtained by sampling as follows: 
    \begin{align}
        \widetilde{x_m} & \sim \mathcal{N}(x_m, (\sigma w)^2) \\
        \widetilde{y_m} & \sim \mathcal{N}(y_m, (\sigma h)^2),
    \end{align}
    
    
    where $w$ and $h$ are the BB width and the height. $\widetilde{x_M}$ and $\widetilde{y_M}$ are calculated similarly from $x_M$ and $y_M$. The parameter $\sigma$ allows to modify the variance of the Gaussian. Figure \ref{fig:sigma} illustrates the effect of $\sigma$ on BB coordinates. By introducing noise in this way, it is still possible to get access to the true identity of each object, which is not possible if we used a detector. Indeed, a detector may miss some hard to detect objects, resulting in a biased analysis.

    Note that adding a Gaussian noise is not sufficient: we have to make sure that the new coordinates are valid (integers such that $0 \leq \widetilde{x_m} < \widetilde{x_M} < \text{W}$ and $0 \leq \widetilde{y_m} < \widetilde{y_M} < \text{H}$, where $W$ and $H$ represent the width and the height of the frame, respectively). We chose the following $\sigma$ parameters: 0, 0.05, 0.1 and 0.2.

%
%
%

    
    \paragraph{Sampling step}
    the second way to simulate noise is by skipping frames: instead of comparing two consecutive frames, we increase the sampling step. Therefore, the visual appearance of objects changes more and this simulates the case when the detector missed some objects or the object was not visible for several frames. We chose the following sampling steps: 1, 2, 4, 8, 16 and 32 frames. Note that this can result in different temporal skip in seconds depending on the video frame rate. This should simply be viewed as gradually including more and more missing detections, making the matching more difficult.
    
    \subsection{Performance measure}
    For a given descriptor-affinity pair, a configuration $\sigma$-step and a sequence of a dataset, we evaluated the average precision for pairs (query object, set of candidate objects) by calculating the ratio of number of correct matches (when we find the same object among the set) over the total number of tested query objects. We reported the mean average precision over sequences on each dataset. 
    
    The configuration ($\sigma=0$, step = 1) refers to the case where the detector can perfectly detect all objects.
    
    \subsection{Implementation details} 
    For HOG, color and grayscale histograms, we used the implementations from OpenCV~\cite{opencv_library}. Models and weights for VGG-19, ResNet-18 and DenseNet-121 come directly from Pytorch~\cite{paszke2019PyTorchImperative}. As for EfficientNet-B0, we used the implementation provided by~\cite{melas-kyriazi2020lukemelasEfficientNetPyTorch}. We relied on pretrained models learned on ImageNet. When using CNN-based models, as recommended by Pytorch, RGB BBs are resized to $224 \times 224$ and normalized.
    OSNet-AIN weights were pretrained by~\cite{zhou2019LearningGeneralisable} on ImageNet and fine-tuned on Market1501~\cite{zheng2015ScalablePerson} and are available in the torchreid library~\cite{zhou2019Torchreidlibrary}. Weights for vehicle ReID were trained by~\cite{wu2018VehicleReIdentification} on VeRi~\cite{liu2016Largescalevehicle, liu2016DeepLearningBased}, CompCars Surveillance~\cite{yang2015LargeScaleCar}, BoxCars~\cite{sochor2018BoxCarsImproving} and unsupervisedly fine-tuned on AI City Challenge dataset~\cite{naphade20182018NVIDIA}.
    
\section{Results and analysis}
\label{results}

 \subsection{General feature performance}
    We summarized all results from the four datasets into four figures to rank the descriptor-affinity pairs according to 24 $\sigma$-step configurations ($\sigma$ and sampling step). For each case and for each dataset, we only reported the best five descriptor-affinity pairs, and for categories of features which were not in the top-5, the best model among them. 
    
\begin{figure*}[!ht]
    \centering               
    \includegraphics[scale = 0.49, trim={0 5 0 20},clip]{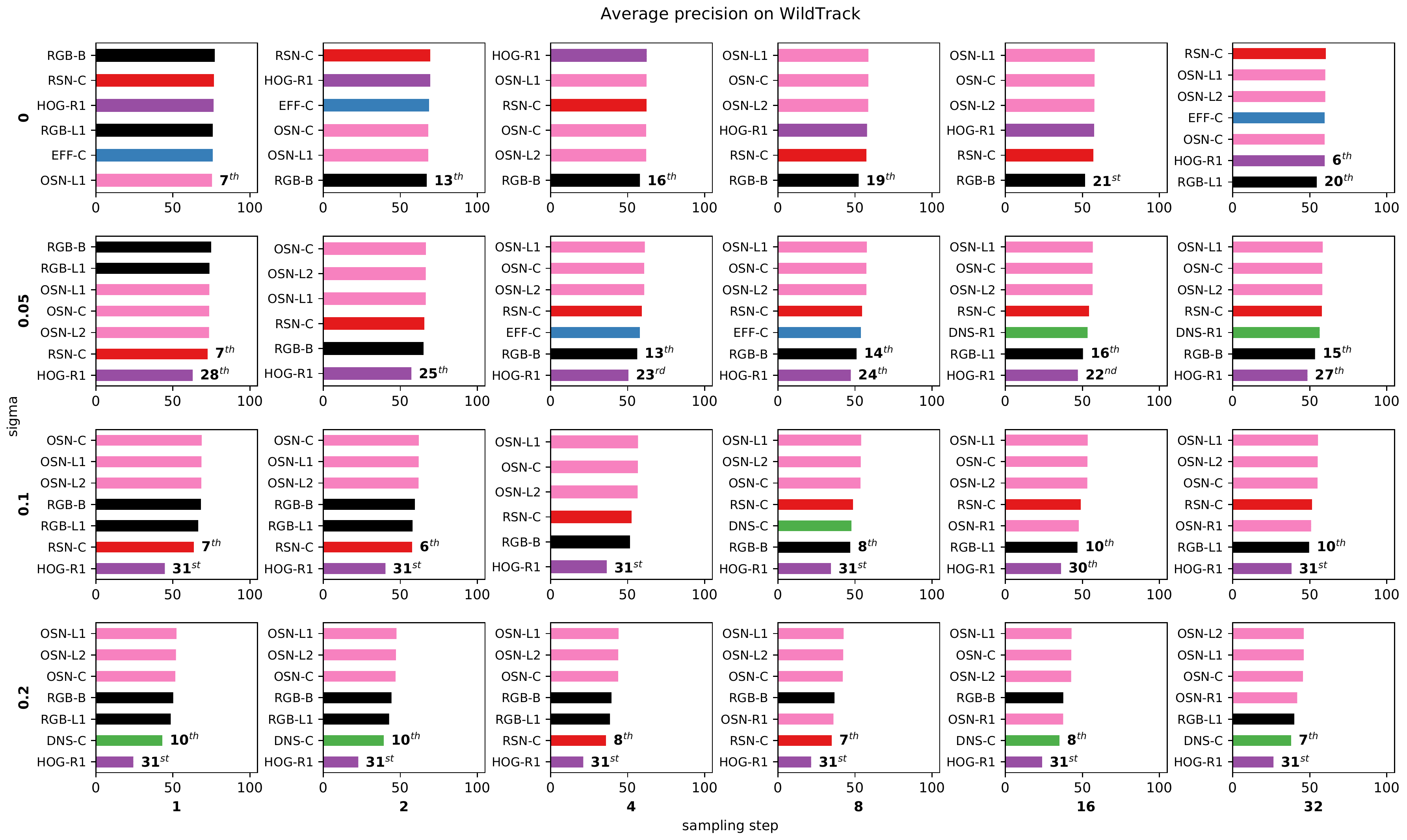}
        \caption{Mean average precision on WildTrack of the five best descriptor-affinity for each configuration $\sigma$-step (when one category of descriptors is not in the top-5, the best result is added). See Tables~\ref{tab:colorDesc} and \ref{tab:hatchAff} for the color and hatching codes used in the figure. Best viewed in color.}
    \label{fig:res_WildTrack}
    \end{figure*}

\begin{figure*}[!ht]
    \centering               
    \includegraphics[scale = 0.49, trim={0 5 0 20},clip]{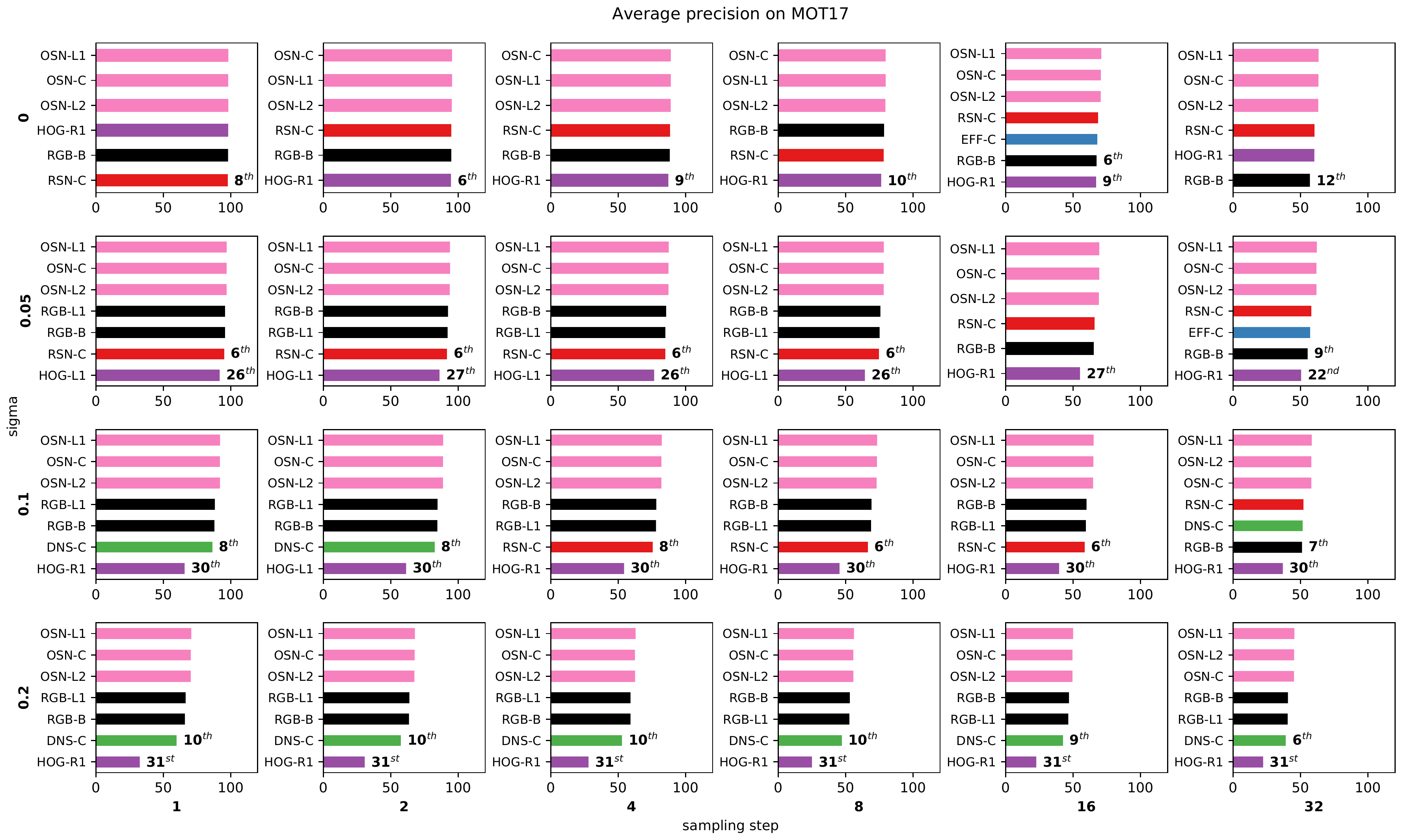}
    \caption{Mean average precision on MOT17 of the five best descriptor-affinity for each configuration $\sigma$-step (when one category of descriptors is not in the top-5, the best result is added). See Tables \ref{tab:colorDesc} and \ref{tab:hatchAff} for the color and hatching codes used in the figure. Best viewed in color.}
    \label{fig:res_MOT20}
    \end{figure*}

\begin{figure*}[!ht]
    \centering               
    \includegraphics[scale = 0.49, trim={0 5 0 20},clip]{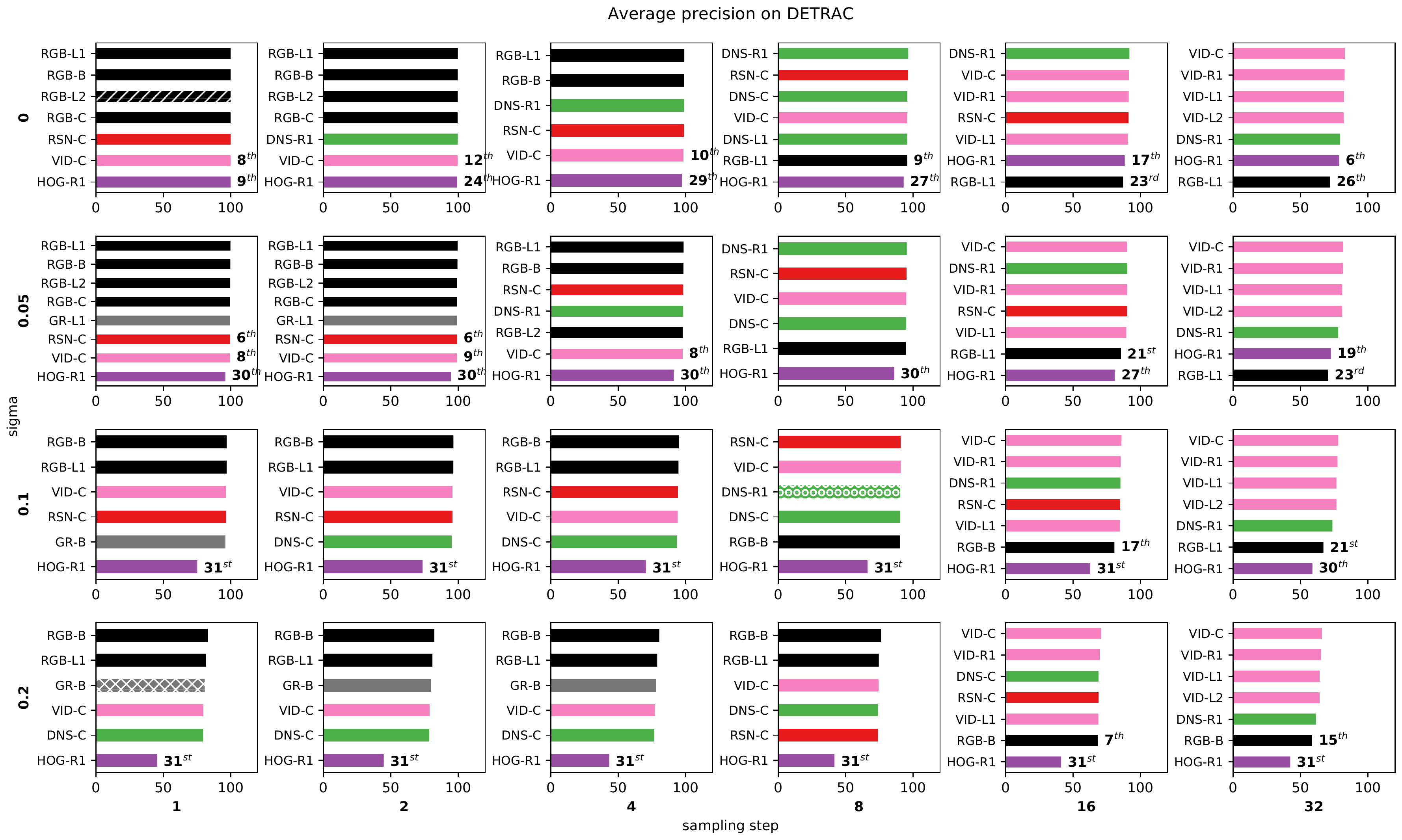}
    \caption{Mean average precision on DETRAC of the five best descriptor-affinity for each configuration $\sigma$-step (when one category of descriptors is not in the top-5, the best result is added). See Tables \ref{tab:colorDesc} and \ref{tab:hatchAff} for the color and hatching codes used in the figure. Best viewed in color.}
    \label{fig:res_DETRAC}
    \end{figure*}

\begin{figure*}[!ht]
    \centering               
    \includegraphics[scale = 0.49, trim={0 5 0 20},clip]{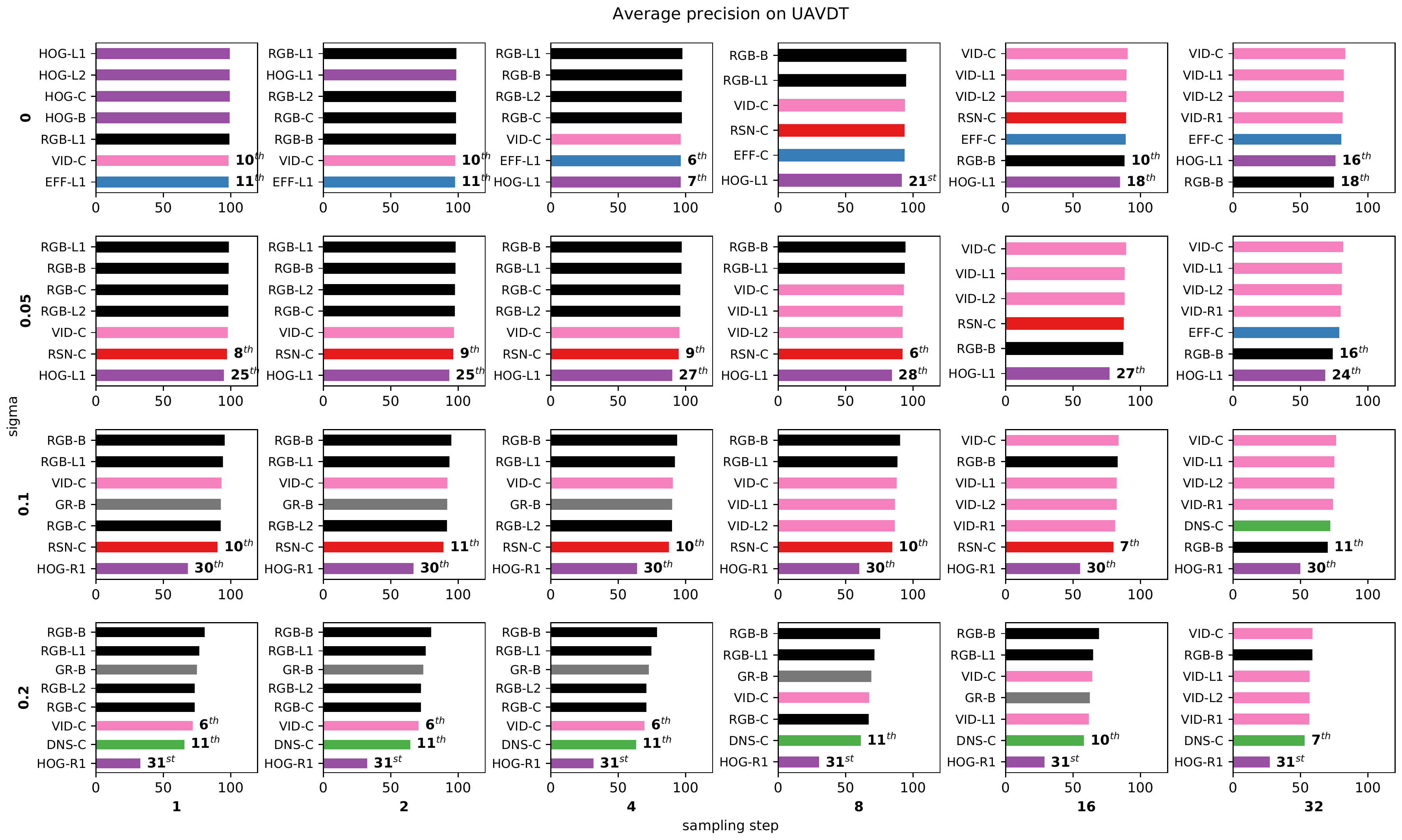}
    \caption{Mean average precision on UAVDT of the five best descriptor-affinity for each configuration $\sigma$-step (when one category of descriptors is not in the top-5, the best result is added). See Tables \ref{tab:colorDesc} and \ref{tab:hatchAff} for the color and hatching codes used in the figure. Best viewed in color.}
    \label{fig:res_UAVDT}
    \end{figure*}

    Tables~\ref{tab:colorDesc} and~\ref{tab:hatchAff} explain the color and hatching codes used in figures~\ref{fig:res_WildTrack}, \ref{fig:res_MOT20}, \ref{fig:res_DETRAC} and~\ref{fig:res_UAVDT}.

    \begin{table}[ht]
    \parbox{.45\linewidth}{
        \caption{Color of the descriptor in figures \ref{fig:res_WildTrack}, \ref{fig:res_MOT20}, \ref{fig:res_DETRAC} and \ref{fig:res_UAVDT}}
        \label{tab:colorDesc}
        \centering
        \begin{tabular}{lc}
        \toprule
            Descriptor                & Color  \\
            \hline
            Color histogram (RGB)     & black  \\
            Grayscale histogram (GR)  & gray   \\
            HOG (HOG)                 & purple \\
            VGG-19 (VGG)              & orange \\
            ResNet-18 (RSN)           & red    \\
            DenseNet-121 (DNS)        & green  \\
            EfficientNet-B0 (EFF)     & blue  \\
            OSNet-AIN (OSN)           & pink \\
            Vehicle ReID (VID)        & pink \\
        \bottomrule
        \end{tabular}
    }
    \hfill
    \parbox{.45\linewidth}{
        \caption{Hatching of the affinity measure in figures~\ref{fig:res_WildTrack}, \ref{fig:res_MOT20}, \ref{fig:res_DETRAC} and \ref{fig:res_UAVDT}}
        \centering
        \begin{tabular}{lc}
            \toprule
            Affinity & Hatching  \\
            \hline
            $\mathrm{L}_1$ (L1)        & \textbackslash\, \textbackslash\, \textbackslash  \\
            $\mathrm{L}_2$ (L2)        & $\mathbin{/}\mathbin{/}\mathbin{/}$   \\
            $C_{\text{rank}1}$ (R1)     & $OO$ \\
            $D_B$ (B)                   & $XXX$ \\
            $S_C$ (C)                   & none \\
        \bottomrule
        \end{tabular}
        \label{tab:hatchAff}
    }
    \end{table}
    
    \subsubsection{$\sigma$-step configuration}
    unsurprisingly, the four figures show that increasing the parameter $\sigma$ and/or the sampling step decreases the matching performance of the best descriptor-affinity model.
    
    \subsubsection{Color and grayscale histograms}
    color histograms are competitive features, in particular for vehicles-centered datasets, for a low sampling step and a low $\sigma$, especially when combined with the Bhattacharyya distance. For these configurations, depending on the datasets, this model is almost always in the top-5. On WildTrack, due to the low framerate, it is not able to discriminate pedestrians when their BBs are separated by more than two seconds, meaning that the objects should not be occluded for too long, or that the detector should have a good recall. We explained this by their color appearance changing too much between these two frames. However, when the BBs do not enclose the object precisely ($\sigma=0.2$), these models ranked almost always in the top-5. This is due to the excessive loss of semantic information when BBs coordinates are imprecise: low-level characteristics such as colors are in that case relevant.
    
    \subsubsection{Histograms of oriented gradients}
    HOG is a good appearance feature descriptor when the BB coordinates correspond to the ground truth. As soon as they get imprecise, the performance of HOG decreases dramatically. 
    Regardless of the dataset, if the BB correspond to the ground truth, the HOG descriptor is among the best models (when it is not in the top-5, the deviation in absolute value from the best model is small). But when $\sigma$ increases, its performance falls on average to the 30$^{th}$ position amongst 35 candidates. In the case of very imprecise BBs, the gap between this feature and others is significant. This is due to the construction of HOG: feature vectors are computed over cells of $8 \times 8$ pixels. So, a small shift in the BB makes this feature non robust.

    
    \subsubsection{CNN-based models}
    this category of models is competitive when the sampling step is high and $\sigma$ moderate. When $\sigma$ is less than $0.1$ and the sampling step over $8$, a CNN-based model is often in the top-5 ranking. 
    Features computed from a VGG-19 descriptor are not competitive against the three other CNN-based models as these features never rank in the top-5. 
    Moreover, cosine similarity is sometimes a good affinity measure but not consistently. 

    \subsubsection{ReID models}
    in the case of pedestrians tracking, OSNet-AIN is generally the best visual feature regardless of the performance of the detector. In almost all configurations, this model ranks first, with either $\mathrm{L}_1$, $\mathrm{L}_2$ distances or cosine similarity. Since this model is trained to discriminate pedestrians, it is made to extract meaningful instance-specific characteristics from images. So, even if the BB of the image is corrupted, it is able to discriminate persons.
    As for vehicles, the model from~\cite{wu2018VehicleReIdentification} ranks in top-5 when $\sigma$ is over 0.1 or when the sampling step is over 4. For BBs more similar to the ground truth, the deviation in absolute value from the best model is low. Cosine similarity is systematically the best affinity measure for this descriptor.
    
 \subsection{Feature performance according to size of objects}
    In addition to two characteristics of the detector (its ability to predict correctly the coordinates of the BBs and to avoid missed detections), the size of objects might influence the choice of the visual feature descriptor. Smaller objects are commonly the hardest targets to track in MOT. But it is unclear how visual features are affected by the size of BBs.
    
\begin{figure}[htbp]
    \centering               
    \includegraphics[scale = 0.5,trim={0 5 0 18},clip]{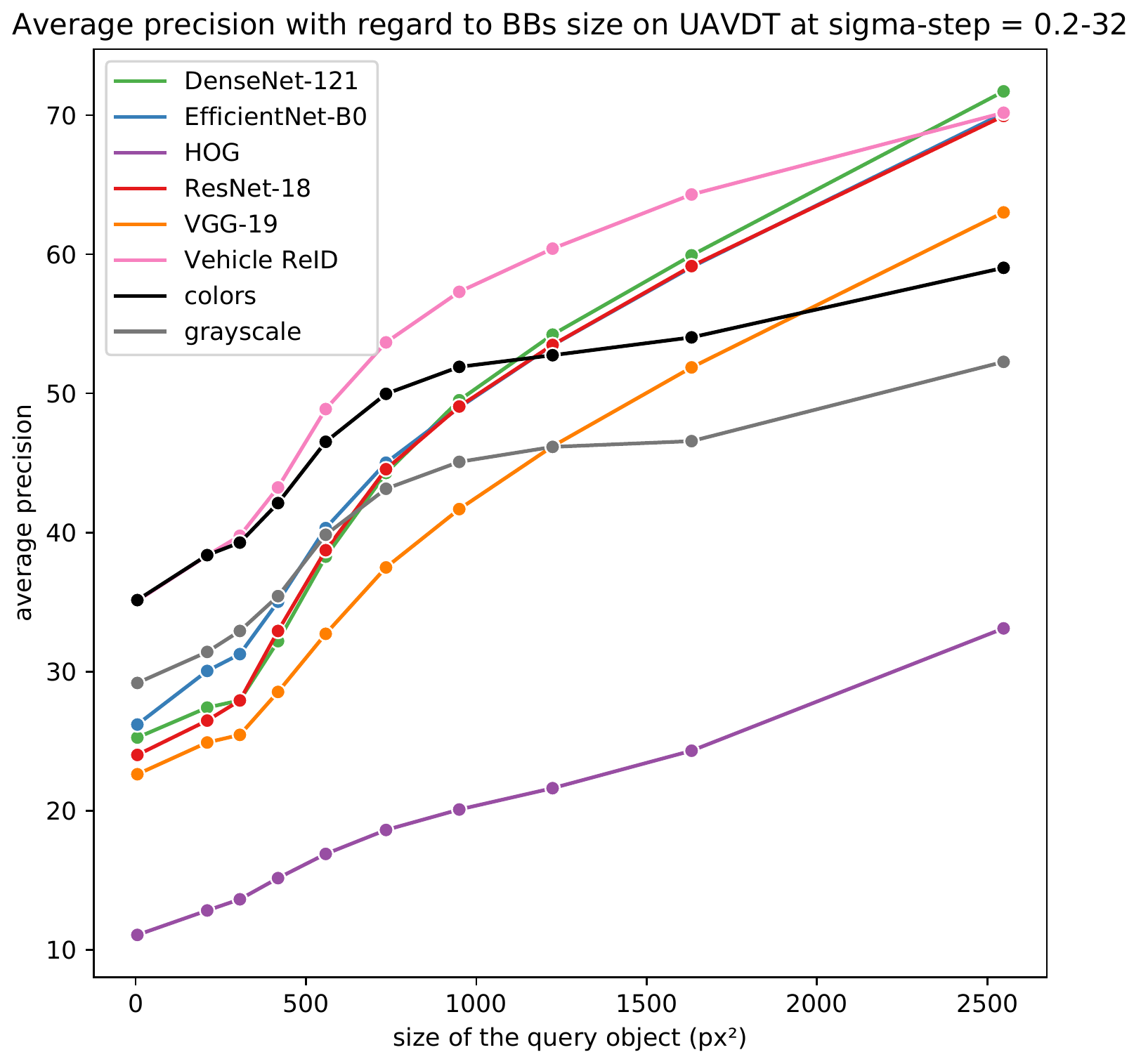}
    \caption{Average precision according to the query object size, with $\sigma = 0.2$, sampling step at 32 and the $\mathrm{L}_2$ distance on UAVDT, computed at each decile. Best viewed in color.}
    \label{fig:res_size}
    \end{figure}

    
   Figure~\ref{fig:res_size} gives the average precision with regard to the query object size, on the UAVDT dataset where there are few occlusions. The configuration $\sigma$-step selected correspond to the hardest one (0.2-32) where differences are more meaningful. For a fair comparison, only the $\mathrm{L}_2$ distance is used.
    
    
    Firstly, for any feature, the larger the query object is, the easier it is to get the correct match. RGB-histograms are among the best visual features for the smallest objects (approximately smaller than 250 pixels$^2$ of area), where it is difficult to extract semantics. But for larger objects, ReID features give the best performance. Then, the tested CNN-based models, except VGG-19 which performs more poorly, yield similar results, but lower than ReID which indicates that performing well on ImageNet does not necessarily produce better features for MOT. Similar conclusions can be drawn from other datasets, with the exception of Wildtrack because of its small scale in terms of available data (cf appendix).

\section{Conclusion}
\label{conclusion}
    In this paper, we compared several feature descriptors in the context of MOT in urban scenes. Our experiments show that features perform differently given the quality of bounding boxes. ReID features, combined with cosine similarity, are one of the best descriptors for pedestrians and vehicles, regardless of the performance of the detector. If these models are not available, color histograms with the Bhattacharyya distance is competitive when the boxes are not too noisy.
    But, as soon as the bounding boxes get noisier, these methods are not able to compete against deep features. Moreover, the size of objects matter on the choice of visual features : in difficult cases, compared to RGB-histograms and modern deep features, ReID features particularly stand out on medium-sized objects. 
    



\section*{Acknowledgment}
We acknowledge the support of the Natural Sciences and Engineering Research Council of Canada (NSERC), [CRDPJ 528786 - 18], [DG 2017-06115], and the support of Arcturus Networks. 




\bibliographystyle{IEEEtran}
\bibliography{references}
%



\newpage
\section*{Appendix}

    In the following, we provide some results on other datasets related on the statements about the effect of the size of an object.

    \begin{figure}[htbp]
    \centering               
    \includegraphics[scale = 0.5,trim={0 5 0 18},clip]{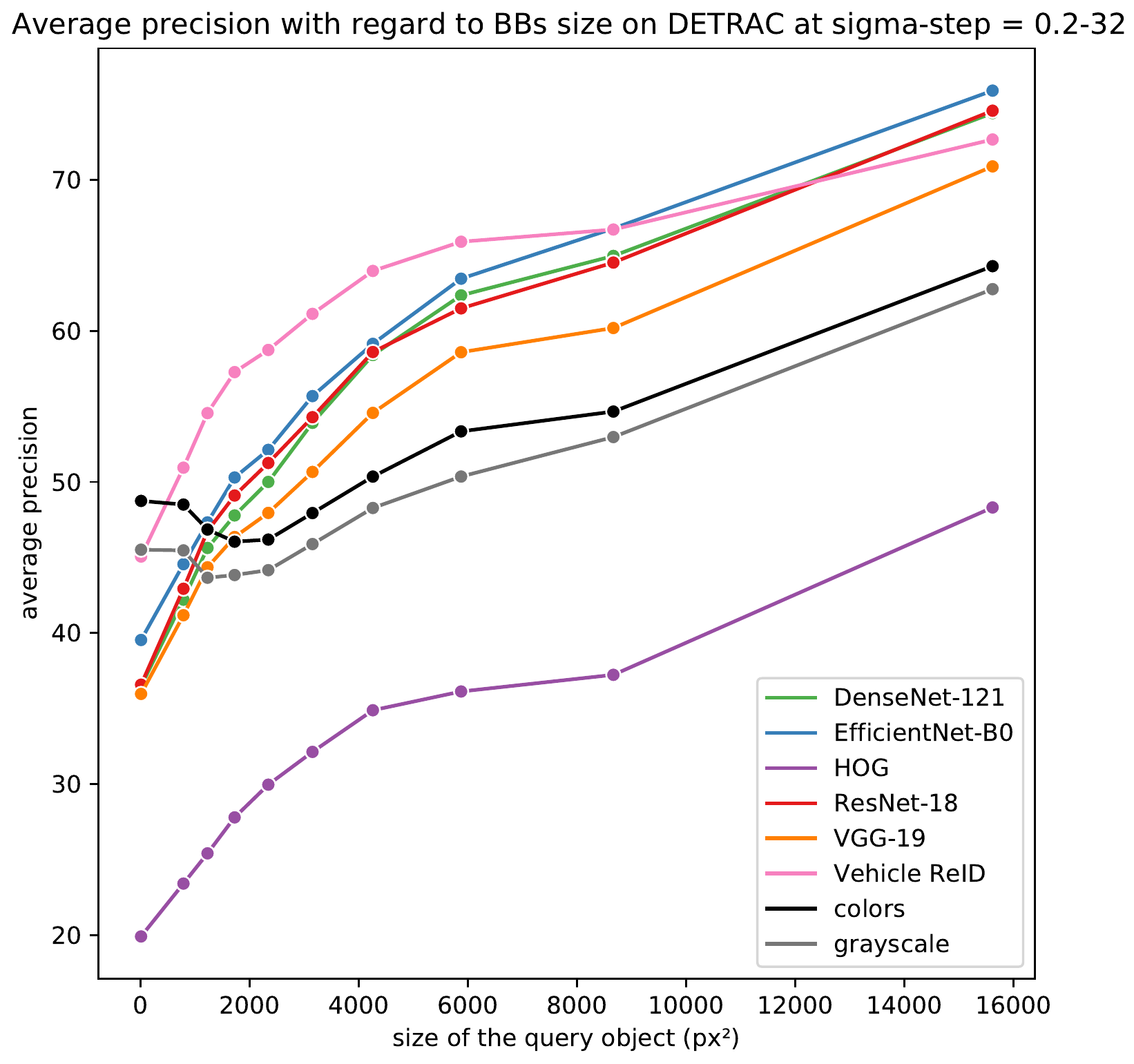}
    \caption{Average precision according to the query object size, with $\sigma = 0.2$, sampling step at 32 and the $\mathrm{L}_2$ distance on DETRAC, computed at each decile. Best viewed in color.}
    \label{fig:DETRAC_absolute_size}
    \end{figure}
    
    \begin{figure}[htbp]
    \centering               
    \includegraphics[scale = 0.5,trim={0 5 0 18},clip]{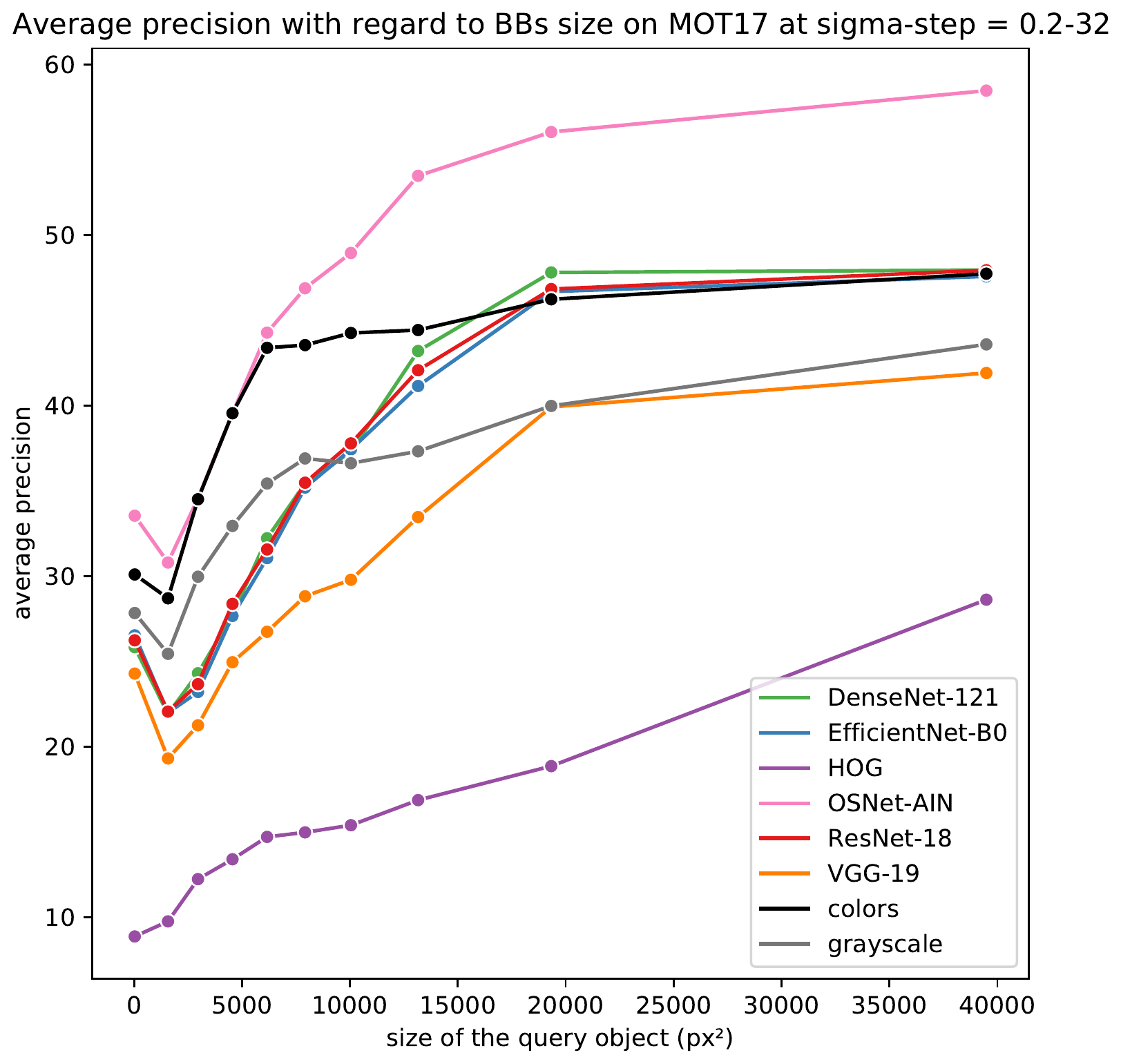}
    \caption{Average precision according to the query object size, with $\sigma = 0.2$, sampling step at 32 and the $\mathrm{L}_2$ distance on MOT17, computed at each decile. Best viewed in color.}
    \label{fig:MOT17_absolute_size}
    \end{figure}
    
    \begin{figure}[htbp]
    \centering               
    \includegraphics[scale = 0.5,trim={0 5 0 18},clip]{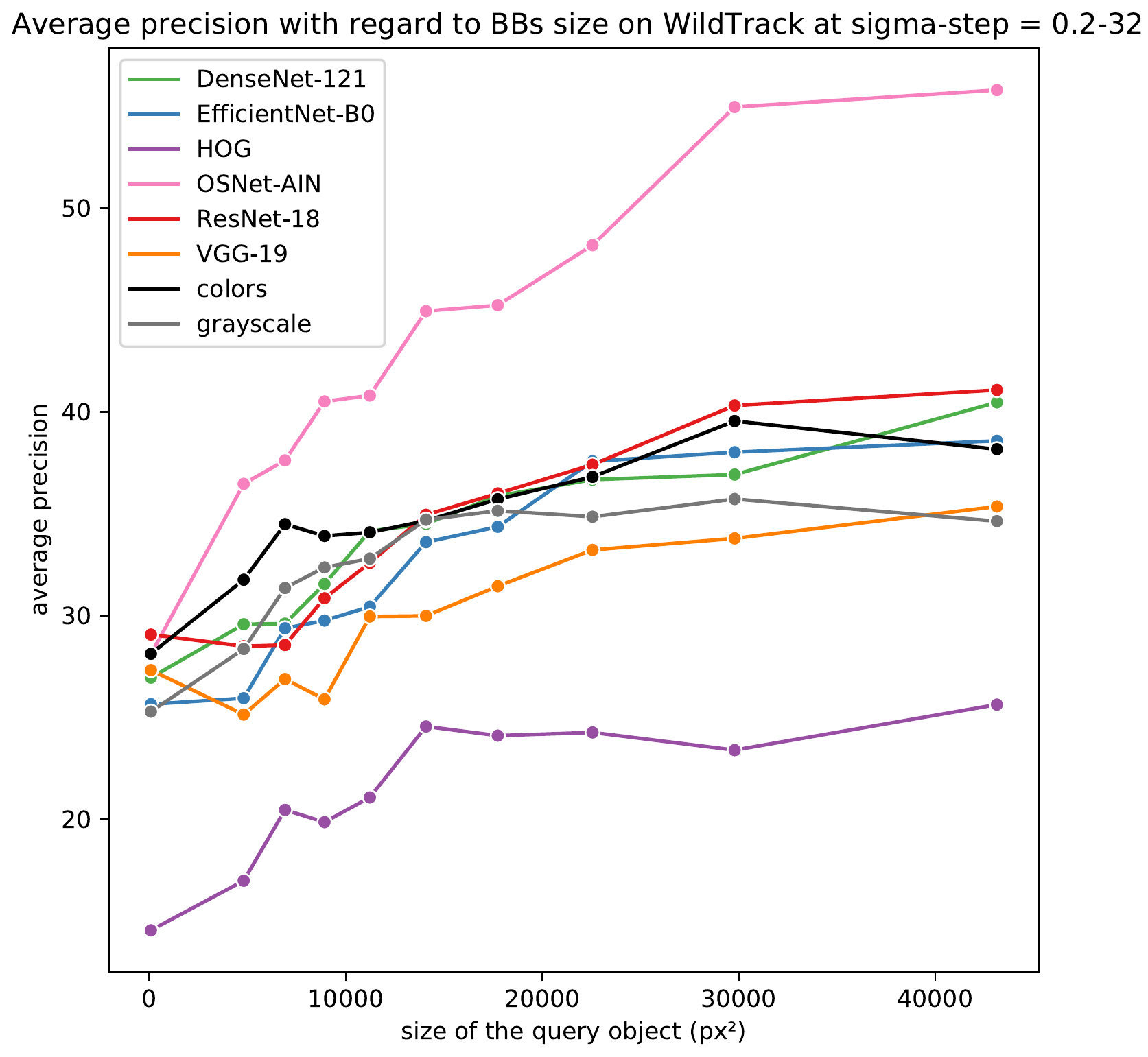}
    \caption{Average precision according to the query object size, with $\sigma = 0.2$, sampling step at 32 and the $\mathrm{L}_2$ distance on WildTrack, computed at each decile. Best viewed in color.}
    \label{fig:WildTrack_absolute_size}
    \end{figure}

\end{document}